\documentclass[conference]{IEEEtran}
\IEEEoverridecommandlockouts
\usepackage{cite}
\usepackage{booktabs}
\usepackage{amsmath,amssymb,amsfonts}
\usepackage{algorithmic}
\usepackage{graphicx}
\usepackage{textcomp}
\usepackage{xcolor}

\usepackage{float}    
\def\BibTeX{{\rm B\kern-.05em{\sc i\kern-.025em b}\kern-.08em
    T\kern-.1667em\lower.7ex\hbox{E}\kern-.125emX}}

\usepackage{url}
\usepackage[hidelinks]{hyperref} 

\begin{document}

\title{Balancing Interpretability and Performance in Motor Imagery EEG Classification: A Comparative Study of ANFIS-FBCSP-PSO and EEGNet\\}

\author{
\IEEEauthorblockN{
Farjana Aktar\textsuperscript{1}, 
Mohd Ruhul Ameen\textsuperscript{1}, 
Akif Islam\textsuperscript{1}, 
Md.~Ekramul Hamid\textsuperscript{1}
}
\IEEEauthorblockA{
\textsuperscript{1}Department of Computer Science and Engineering, University of Rajshahi\\
Rajshahi 6205, Bangladesh\\
farjana.aktar.cseru@gmail.com, 
ameensunny242@ru.ac.bd, 
iamakifislam@gmail.com, 
ekram\_hamid@ru.ac.bd
}
}
\maketitle


\vspace{0.5cm}
\begin{center}
\small Accepted at the 2026 IEEE 2nd International Conference on Quantum Photonics, Artificial Intelligence and Networking (QPAIN 2026). Copyright IEEE.
\end{center}
\vspace{0.5cm}

\begin{abstract}
Achieving both accurate and interpretable classification of motor-imagery EEG remains a key challenge in brain-computer interface (BCI) research. In this paper, we compare a transparent fuzzy-reasoning approach (ANFIS--FBCSP--PSO) with a well-known deep-learning benchmark (EEGNet) using the publicly available BCI Competition IV-2a dataset. The ANFIS pipeline combines filter-bank common spatial pattern feature extraction with fuzzy IF--THEN rules optimized via particle-swarm optimization, while EEGNet learns hierarchical spatial-temporal representations directly from raw EEG data. In within-subject experiments, the fuzzy-neural model performed better (68.58\% $\pm$ 13.76\% accuracy, $\kappa$ = 58.04\% $\pm$ 18.43), while in cross-subject (LOSO) tests, the deep model exhibited stronger generalization (68.20\% $\pm$ 12.13\% accuracy, $\kappa$ = 57.33\% $\pm$ 16.22). The study therefore provides practical guidance for selecting MI-BCI systems according to the design goal: interpretability or robustness across users. Future investigations into transformer-based and hybrid neuro-symbolic frameworks are expected to further advance transparent EEG decoding.
\end{abstract}


\section{Introduction}
Brain--Computer Interfaces (BCIs) enable direct communication between neural activity and external devices, supporting assistive technologies for people with motor impairments and expanding human--computer interaction~\cite{nicolas2012brain, he2019practical, lawhern2018eegnet}. Among non-invasive modalities, Electroencephalography (EEG) remains the most practical choice due to its high temporal resolution, portability, and low cost~\cite{lotte2018review}. A widely studied EEG-BCI paradigm is Motor Imagery (MI), where users mentally rehearse movements without execution, producing characteristic neural patterns that can be decoded into control commands. Despite steady progress, designing MI-BCI systems that are both accurate and interpretable in real-time settings remains challenging.

Classical MI-EEG pipelines typically rely on feature engineering. After preprocessing to reduce noise and artifacts (e.g., via Independent Component Analysis)~\cite{hyvarinen2000independent}, spatial--spectral features are extracted using methods such as Filter Bank Common Spatial Pattern (FBCSP)~\cite{ang2012filter}. By decomposing signals into frequency sub-bands and learning discriminative spatial filters within each band, FBCSP often yields strong subject-specific representations. However, these handcrafted pipelines can require heuristic tuning and provide limited transparency about how features map to motor tasks.

Deep learning approaches address feature design by learning representations directly from EEG. Convolutional models such as EEGNet and DeepConvNet capture relevant temporal and spatial structure from raw signals~\cite{schirrmeister2017deep, lawhern2018eegnet}, and EEGNet in particular offers a compact architecture suitable for small datasets and real-time use. Despite their strong performance, many deep learning approaches lack transparency, limiting their acceptance in clinical environments.

Explainable Artificial Intelligence (XAI) aims to bridge this gap by providing transparent and accountable decision mechanisms~\cite{xu2020explainable}. Rather than trying to explain a complex model after the fact, inherently interpretable approaches are transparent from the start. Adaptive Neuro-Fuzzy Inference Systems (ANFIS) merge neural learning with fuzzy IF–THEN rules, producing decisions that can be understood in clear, physiologically meaningful terms for MI-EEG classification~\cite{jang1993anfis}. Their performance can be strengthened using Particle Swarm Optimization (PSO) to tune model parameters and guide feature selection, improving accuracy while preserving interpretability.

In this study, we examine the practical trade-off between interpretability and performance in MI-EEG decoding on the BCI Competition IV-2a dataset. We compare an optimized ANFIS--FBCSP--PSO pipeline that yields explicit fuzzy rules with EEGNet, a strong end-to-end deep learning baseline. Both models are evaluated under within-subject and leave-one-subject-out (LOSO) protocols using accuracy, F1-score and Cohen's $\kappa$ metrics, alongside qualitative analysis of interpretability via extracted rule sets. Our main contributions are:
\begin{enumerate}
    \item \textbf{Dual-model evaluation:} A controlled comparison between an interpretable bio-inspired approach (ANFIS--FBCSP--PSO) and a deep learning benchmark (EEGNet) under consistent preprocessing and evaluation settings on BCI Competition IV-2a.
    \item \textbf{Interpretability--generalization trade-off:} Empirical analysis of how rule-based interpretability and cross-subject robustness differ across within-subject and LOSO protocols.
    \item \textbf{Actionable guidance:} Practical recommendations for selecting architectures based on application needs, ranging from personalized explainable BCIs to scalable user-general models.
\end{enumerate}

\section{Related Works}

The BCI Competition IV-2a (BCICIV-2a) dataset serves as a widely recognized benchmark for evaluating motor imagery (MI) EEG classification methods. It includes EEG recordings from nine participants, each performing four distinct motor imagery tasks. Early research primarily focused on handcrafted feature extraction techniques, such as Filter Bank Common Spatial Pattern (FBCSP), which were often combined with convolutional neural networks (CNNs) to improve classification performance. For instance, Sakhavi et al.~\cite{sakhavi2018fbcspcnn} introduced a hybrid FBCSP CNN framework that achieved an accuracy of 78.1\%, though its ability to generalize across different subjects remained limited.

Subsequent developments in end-to-end deep learning led to architectures like DeepConvNet, ShallowConvNet~\cite{schirrmeister2017deep}, and EEGNet~\cite{lawhern2018eegnet}, which eliminated the need for manual feature design. These models achieved improved classification accuracy reaching approximately 80.3\% and enhanced computational efficiency. However, their performance often degraded in the presence of noisy and non-stationary EEG signals, a persistent challenge in real-world BCI applications.

More recently, researchers have incorporated attention mechanisms to better capture the temporal dynamics and contextual relationships present in EEG signals. Architectures such as ATCNet~\cite{altaheri2023atcnet} and EEG-Conformer~\cite{song2023eegconformer} reflect this evolution, combining convolutional layers with transformer-based components to achieve reported accuracies of up to 81.98\%. However, these performance gains typically come with increased computational complexity and greater hardware demands, which may limit their practicality in real-world BCI deployments.

Parallel to deep learning advances, bio-inspired models particularly Adaptive Neuro-Fuzzy Inference Systems (ANFIS) have gained attention for their ability to manage EEG signal uncertainty and nonlinearity while preserving model interpretability. Hybrid frameworks such as Deep Fuzzy Neural Networks (DFNN)~\cite{nguyen2023hybrid} and swarm-optimized CNN fuzzy systems~\cite{zhao2024bioinspired} have achieved accuracies of up to 86.2\%, underscoring the advantages of combining fuzzy reasoning with deep feature representation.

Despite these advancements, a consistent trade-off remains between accuracy and generalization. In this context, the present study investigates a bio-inspired hybrid pipeline (ANFIS--FBCSP--PSO) and compares it against EEGNet to examine the balance between interpretability and cross-subject robustness in MI-EEG decoding using the BCICIV-2a dataset.

\begin{figure*}[!t]
    \centering
    \includegraphics[width=0.8\textwidth]{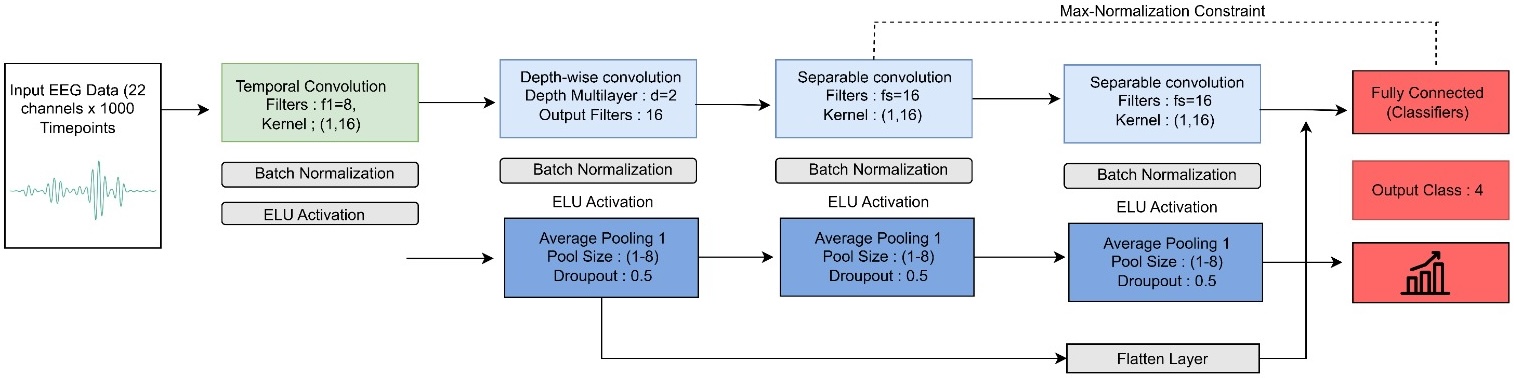}
    \caption{The EEGNet Architecture for Motor Imagery dataset. A compact and generalizable Convolutional Neural Network (CNN) for EEG classification, featuring Temporal, Depthwise, and Separable convolution blocks optimized for extracting frequency, spatial, and temporal features from raw EEG time series data.}
    \label{fig:eegnet_architecture}
\end{figure*}

\section{Methodology}

\begin{figure}[!t]
    \centering
    \includegraphics[width=0.9\linewidth]{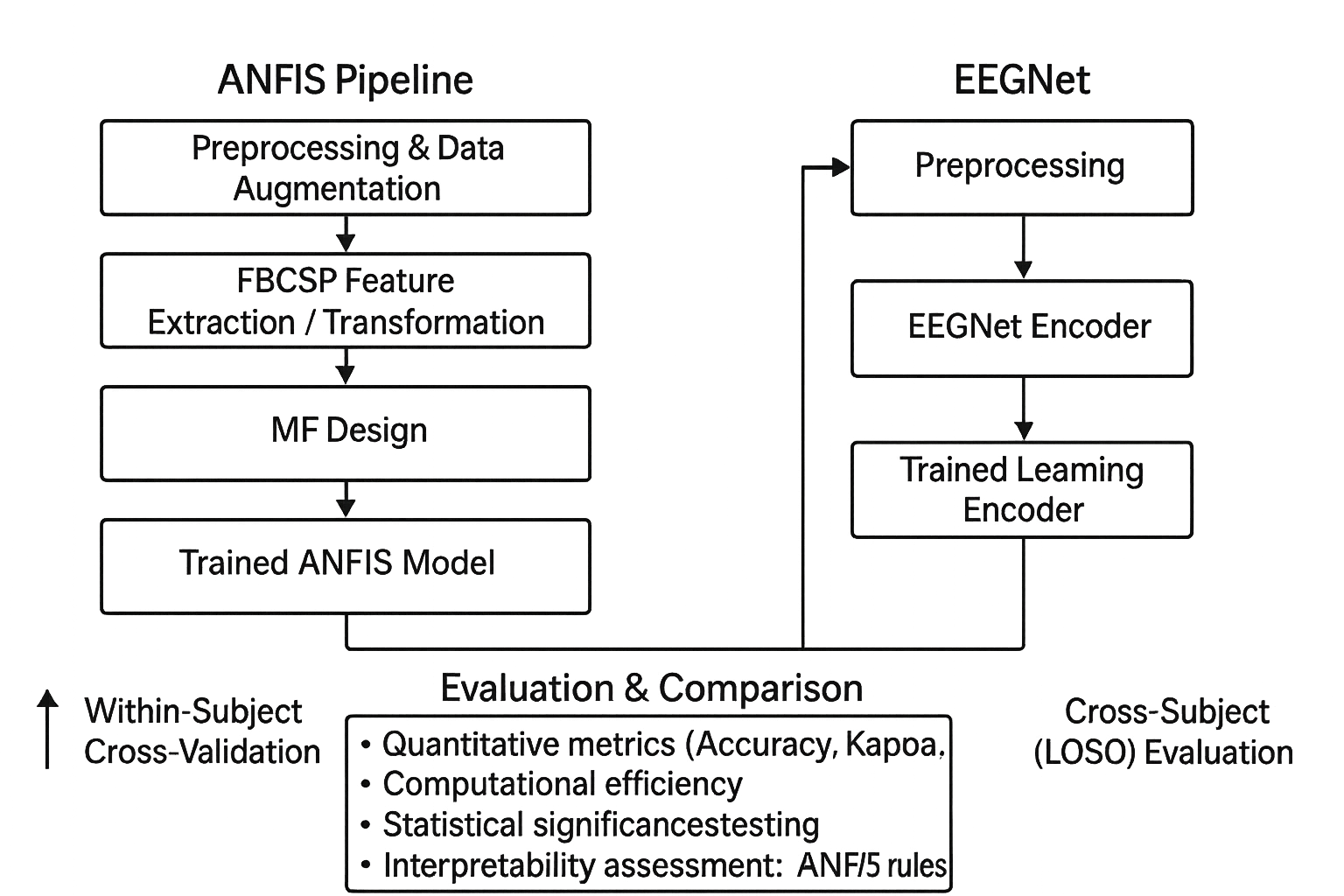}
    \caption{Our Proposed Methodology for EEG Classification on BCI IV 2A dataset}
    \label{fig:method_architecture}
\end{figure}

To assess the performance of the two classification approaches, experiments were conducted using a well-established benchmark dataset — the BCI Competition IV-2a (BCI IV-2a)\cite{bbci_iv_2a} made publicly available by the Graz University of Technology. The subsequent subsections detail the dataset characteristics, preprocessing procedures, data augmentation strategy, and the model architectures employed for motor imagery (MI) EEG classification, as illustrated in Figure~\ref{fig:method_architecture}. 

\subsection{Dataset Description}

The \textit{BCI Competition IV-2a (BCI IV-2a)} dataset~\cite{bbci_iv_2a} is widely recognized as a benchmark in motor imagery (MI) EEG research owing to its comprehensive and high quality recordings. The dataset comprises electroencephalography (EEG) data collected from nine healthy subjects, each performing four distinct MI tasks corresponding to the movements of the \textit{left hand}, \textit{right hand}, \textit{both feet}, and \textit{tongue}. Each participant completed two recording sessions, with a total of 288 trials per session (72 trials per class).  

EEG signals were acquired using 22 Ag/AgCl electrodes positioned according to the international 10--20 system, sampled at a frequency of 250~Hz. The signals were preprocessed using a band-pass filter ranging from 0.5~Hz to 100~Hz to retain relevant frequency components and a notch filter at 50~Hz to mitigate power line interference.  

For subsequent analysis, only the 0--4~second interval following the presentation of the visual cue was utilized, as this period captures the most significant event-related desynchronization/synchronization (ERD/ERS) patterns linked to motor imagery activity. Consequently, each trial is represented as a matrix of size $(22 \times 1000)$, corresponding to 22 EEG channels and 1000 temporal samples. In total, the dataset provides 5,184 trials across all subjects, enabling both subject-specific and cross-subject classification experiments.

\subsection{Preprocessing}

Prior to model training, the raw EEG recordings were standardized to ensure uniform feature scaling across all trials, thereby enhancing training stability and convergence. Specifically, Z-score normalization was applied to each trial \(X \in \mathbb{R}^{C \times T}\), and the normalized signal \(\hat{X}\) was computed as follows:

\begin{equation}
\hat{X} = \frac{X - \mu_X}{\sigma_X}
\end{equation}

where \(\mu_X\) and \(\sigma_X\) denote the mean and standard deviation of the corresponding trial, respectively. After normalization, each processed trial \(\hat{X}^i\) was assigned to its respective motor imagery label \(y_i \in \{1,2,3,4\}\).

To further improve data quality, Independent Component Analysis (ICA) was employed for artifact removal, effectively suppressing ocular and muscular artifacts detected in the EEG signals. In addition, all EEG channels were standardized within each recording session to minimize amplitude discrepancies across trials. The resulting clean and normalized EEG trials were subsequently used as inputs for both deep learning and bio-inspired classification pipelines.

\subsection{Data Augmentation}

Motor imagery EEG datasets typically contain a limited number of samples, which increases the risk of overfitting in deep learning models. To address this issue, a segmentation recombination (S\&R) augmentation approach was applied~\cite{lotte2015signal}. The augmentation strategy was applied only to the training data in both within-subject and LOSO settings. This method enhances data diversity by creating synthetic trials through the recombination of temporal segments drawn from different trials within the same MI class.

Formally, let 
\begin{equation}
T = \{X_1, X_2, \dots, X_M\}
\end{equation}
denote the set of all training trials corresponding to a given class. Each trial is partitioned into \(K\) non-overlapping temporal segments as follows:

\begin{equation}
X_i = \big[ X_i(1), X_i(2), \dots, X_i(K) \big], \quad i = 1, \dots, M
\end{equation}

New artificial trials are generated by randomly selecting one segment from different trials for each temporal position:

\begin{equation}
\tilde{X}_j = \big[ X_{R_1}(1), X_{R_2}(2), \dots, X_{R_K}(K) \big]
\end{equation}

where \(R_k\) is a randomly chosen index from the set \(\{1, \dots, M\}\). This augmentation strategy maintains the temporal dependencies of the original EEG signals while increasing sample variability. Consequently, it improves the generalization capability of both deep learning architectures and ANFIS-based models, leading to more robust motor imagery classification.

\subsection{Model Architectures}

\subsubsection{EEGNet}

EEGNet is a lightweight convolutional neural network specifically designed for MI EEG classification. It takes a single-trial EEG segment of shape $(C, T)$, reshaped to $(B, 1, C, T)$ for batch processing shown in Figure~\ref{fig:eegnet_architecture}. The network first applies temporal convolution to extract frequency-specific patterns, followed by depthwise spatial convolution to learn spatial filters across channels. Average pooling and dropout layers reduce dimensionality and prevent overfitting. A separable convolution captures higher level temporal spatial interactions, followed by another pooling and dropout layer. The flattened features are then passed to a fully connected layer to predict MI classes. The hyperparameters for EEGNet model shown in Table~\ref{tab:hyperparameters_eeg}.

\begin{table}[!t]
\caption{EEGNet Hyperparameters for the BCI IV-2a Dataset}
\label{tab:hyperparameters_eeg}
\centering
\begin{tabular}{ll}
\toprule
\textbf{Layer} & \textbf{Hyperparameter} \\
\midrule
Temporal Convolution & Filters $f_1 = 8$, Kernel Size $= 64$ \\
Depthwise Convolution & Depth Multiplier $D = 2$ \\
Pooling & $P_1 = P_2 = 8$ \\
Separable Convolution & Filters $f_2 = f_1 \times D = 16$, Kernel Size $= 16$ \\
Dropout & $p = 0.5$ \\
Fully Connected Output & $N = 4$ Classes \\
\bottomrule
\end{tabular}
\end{table}

\subsubsection{ANFIS--FBCSP--PSO Architecture}

The ANFIS-based pipeline integrates Filter Bank Common Spatial Pattern (FBCSP) feature extraction with ANFIS optimized via Particle Swarm Optimization (PSO). Preprocessed EEG trials are filtered into multiple frequency bands (Theta, Mu, Low/Mid/High Beta, and Mu+Beta). Common Spatial Pattern (CSP) features are computed for each band, and the most discriminative features are selected.

\begin{figure*}[!t]
    \centering
    \includegraphics[width=0.5\textwidth]{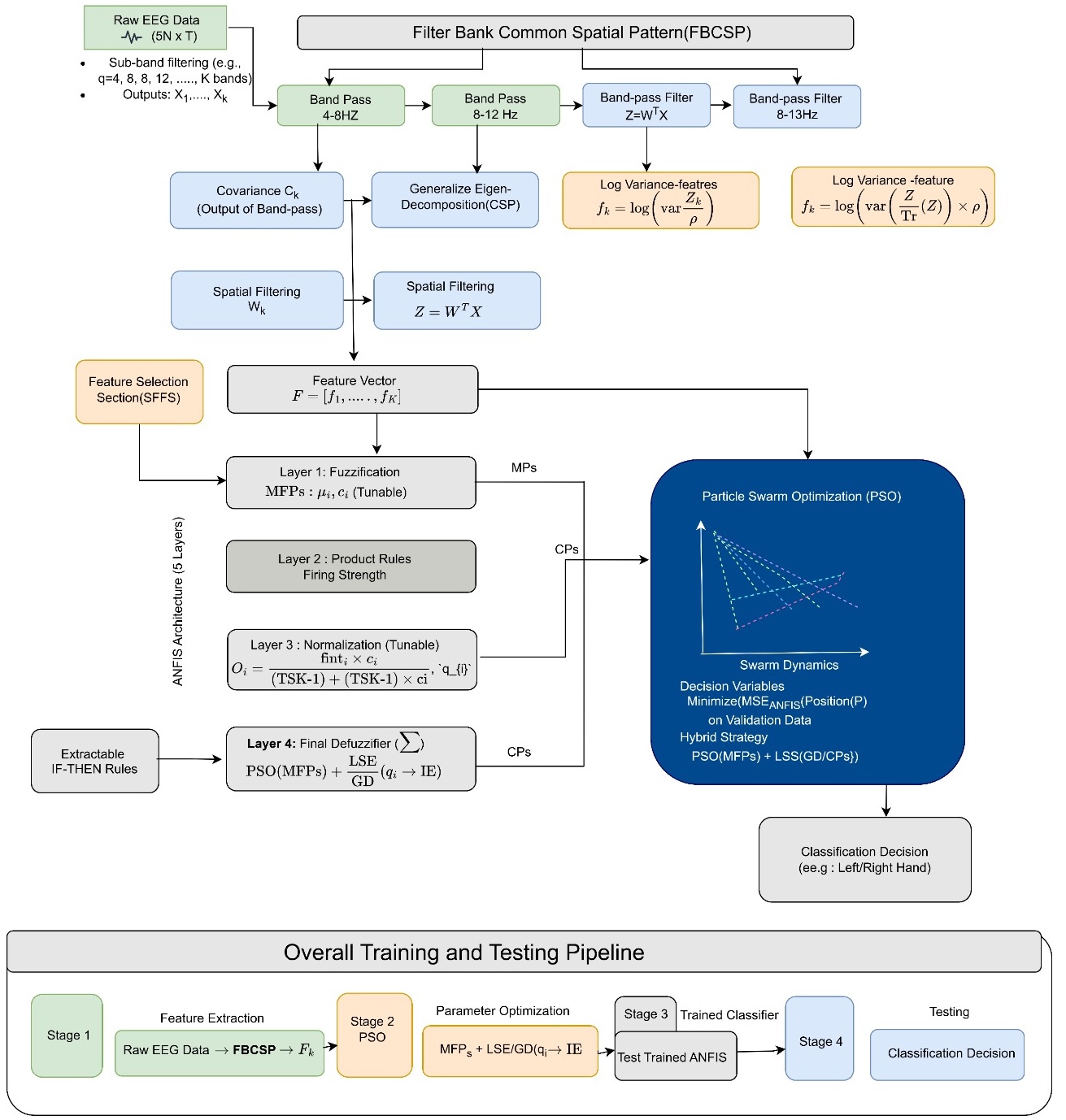}
    \caption{The ANFIS-FBCSP-PSO Hybrid Architecture. This model integrates the FBCSP module for robust EEG feature extraction, an ANFIS for interpretable, rule-based classification, and Particle Swarm Optimization (PSO) for the hybrid tuning of the ANFIS parameters.}
    \label{fig:anfis_architecture}
\end{figure*}

PSO is used to optimize ANFIS parameters, including membership function shapes, widths, and rule weights, with the fitness function defined as classification accuracy on a validation set. The ANFIS layer comprises five layers: fuzzification, rule firing strength computation, normalization, weighted linear combination, and aggregation to produce predicted MI class labels shown in Figure~\ref{fig:anfis_architecture}. The hyperparameters for this model shown in Table~\ref{tab:hyperparameters}.

\begin{table}[!t]
\caption{Hyperparameters for the ANFIS–FBCSP–PSO Model}
\label{tab:hyperparameters}
\centering
\begin{tabular}{p{1.2cm} p{6.2cm}}
\toprule
\textbf{Component} & \textbf{Hyperparameters / Values} \\
\midrule
FBCSP & CSP components per band: 4; Selected features: 4; Frequency bands: Theta (4--8 Hz), Mu (8--12 Hz), Low Beta (12--16 Hz), Mid Beta (16--20 Hz), High Beta (20--24 Hz), Beta (24--30 Hz), Mu+Beta (8--30 Hz); Bandpass filter: 5th-order Butterworth. \\[3pt]

PSO & Particles: 30--50; Iterations: 50--100; Cognitive coefficient $c_1$: 1.5--2.0; Social coefficient $c_2$: 1.5--2.0; Inertia weight $w$: 0.7--1.0; Fitness function: validation accuracy. \\[3pt]

ANFIS & Inputs: selected FBCSP features; Membership functions per input: 2--3; Type: Gaussian/Bell/Triangular; Number of rules: combination of MFs; Epochs: 100--300; Learning rate: 0.01--0.05. \\
\bottomrule
\end{tabular}
\end{table}

This integrated framework combines multi-band spatial features, fuzzy reasoning, and PSO optimization to achieve accurate and robust MI-EEG classification.

\subsection{Quantitative Evaluation Metrics}

The performance of the developed models was assessed using several quantitative metrics to ensure a comprehensive evaluation:

\begin{itemize}
    \item \textbf{Accuracy:} Represents the ratio of correctly classified trials to the total number of trials, serving as a basic indicator of overall model performance.
    \item \textbf{Cohen’s Kappa ($\kappa$):} A statistical measure that adjusts for the possibility of random agreement, providing a more dependable evaluation of consistency in multi-class classification tasks.
    \item \textbf{F1-Score:} Calculated as the harmonic mean of precision and recall, this metric is crucial for determining the balance between correctly identified and missed instances across different classes.

\end{itemize}

\section{Results and Discussion}

Two complementary evaluation strategies were employed to analyze the model performance. In the \textit{within-subject} evaluation, each participant’s dataset was divided into 80\% for training and 20\% for validation. This method examines the model’s capability to recognize motor imagery EEG patterns when trained and tested on data from the same individual, thereby reflecting subject-specific learning.

In contrast, the \textit{cross-subject} evaluation (Leave-One-Subject-Out, LOSO) involved training the model on data from all participants except one, which was then used exclusively for testing. This approach evaluates the model’s ability to generalize across different individuals. The key performance indicators analyzed included Accuracy, Precision, Recall, F1-Score, and Cohen’s Kappa ($\kappa$).

\subsection{Results}

\subsubsection{Within-Subject Analysis}

The detailed within-subject results for the EEGNet and ANFIS-FBCSP-PSO models are presented in Tables III and IV, respectively. On average, EEGNet achieved an accuracy of $63.79\% \pm 8.49$ with a Kappa score of $51.54 \pm 11.67$, while ANFIS-FBCSP-PSO obtained an accuracy of $68.58\% \pm 13.76$ and a Kappa value of $58.04 \pm 18.43$. These findings demonstrate that the ANFIS-FBCSP-PSO model is more effective in learning subject-specific discriminative patterns, resulting in stronger agreement between predicted and actual class labels.

\begin{table}[!t]
\caption{Within-Subject Evaluation Results for EEGNet}
\label{tab:eegnet_within}
\centering
\footnotesize
\setlength{\tabcolsep}{3pt} 
\begin{tabular}{lccccc}
\toprule
\textbf{Subj.} & \textbf{Acc.} & \textbf{Prec.} & \textbf{Rec.} & \textbf{F1} & \textbf{Kappa} \\
\midrule
S1 & 63.79 & 63.96 & 67.26 & 62.78 & 52.10 \\
S2 & 46.55 & 49.46 & 45.85 & 46.07 & 28.17 \\
S3 & 63.79 & 63.01 & 61.95 & 60.33 & 51.10 \\
S4 & 55.17 & 57.29 & 51.88 & 48.35 & 38.70 \\
S5 & 63.79 & 65.76 & 67.85 & 63.95 & 52.35 \\
S6 & 65.52 & 65.80 & 66.41 & 65.39 & 54.04 \\
S7 & 70.69 & 71.63 & 72.18 & 70.48 & 61.10 \\
S8 & 70.69 & 71.52 & 72.57 & 69.74 & 61.03 \\
S9 & 74.14 & 75.87 & 73.32 & 72.82 & 65.27 \\
\midrule
\textbf{Mean} & \textbf{63.79} & \textbf{64.92} & \textbf{64.37} & \textbf{62.21} & \textbf{51.54} \\
\textbf{Std}  & \textbf{8.49}  & \textbf{7.99}  & \textbf{9.60}  & \textbf{9.40}  & \textbf{11.67} \\
\bottomrule
\end{tabular}
\end{table}

\begin{table}[!t]
\caption{Within-Subject Evaluation Results for ANFIS–FBCSP–PSO}
\label{tab:anfis_within}
\centering
\footnotesize
\setlength{\tabcolsep}{3pt} 
\begin{tabular}{lccccc}
\toprule
\textbf{Subj.} & \textbf{Acc.} & \textbf{Prec.} & \textbf{Rec.} & \textbf{F1} & \textbf{Kappa} \\
\midrule
S1 & 70.69 & 74.87 & 75.45 & 70.71 & 61.41 \\
S2 & 62.07 & 61.57 & 64.51 & 60.86 & 49.36 \\
S3 & 74.13 & 75.14 & 74.76 & 74.48 & 65.56 \\
S4 & 60.34 & 62.50 & 60.36 & 60.75 & 47.23 \\
S5 & 48.27 & 50.92 & 48.14 & 48.55 & 30.84 \\
S6 & 55.17 & 56.29 & 54.51 & 54.43 & 39.58 \\
S7 & 91.37 & 91.52 & 91.87 & 91.40 & 88.50 \\
S8 & 84.48 & 83.88 & 83.72 & 83.35 & 79.07 \\
S9 & 70.69 & 71.89 & 70.98 & 70.84 & 60.84 \\
\midrule
\textbf{Mean} & \textbf{68.58} & \textbf{69.84} & \textbf{69.37} & \textbf{68.38} & \textbf{58.04} \\
\textbf{Std}  & \textbf{13.76} & \textbf{13.17} & \textbf{13.95} & \textbf{13.72} & \textbf{18.43} \\
\bottomrule
\end{tabular}
\end{table}

\subsection{Cross-Subject Analysis}

Tables~\ref{tab:eegnet_cross} and~\ref{tab:anfis_cross} present the cross-subject performance outcomes for the EEGNet and ANFIS-FBCSP-PSO models. On average, EEGNet achieved an accuracy of 68.20\% ± 12.13\% with a Kappa value of 57.33\% ± 16.22\%, whereas ANFIS-FBCSP-PSO attained an accuracy of 65.71\% ± 14.89\% and a Kappa score of 53.66 ± 20.52. These findings indicate that EEGNet demonstrates stronger generalization capabilities when tested on unseen subjects.

\begin{table}[!t]
\caption{Cross-Subject Evaluation Results for EEGNet}
\label{tab:eegnet_cross}
\centering
\footnotesize
\setlength{\tabcolsep}{3pt} 
\begin{tabular}{lccccc}
\toprule
\textbf{Subj.} & \textbf{Acc.} & \textbf{Prec.} & \textbf{Rec.} & \textbf{F1} & \textbf{Kappa} \\
\midrule
S1 & 68.96 & 70.58 & 68.95 & 69.30 & 58.24 \\
S2 & 60.34 & 59.02 & 58.53 & 56.04 & 46.51 \\
S3 & 65.52 & 70.56 & 67.02 & 65.10 & 54.38 \\
S4 & 51.72 & 52.97 & 55.28 & 52.99 & 35.09 \\
S5 & 63.79 & 65.53 & 63.14 & 62.32 & 50.27 \\
S6 & 56.89 & 63.16 & 59.19 & 57.02 & 43.18 \\
S7 & 87.93 & 86.74 & 87.17 & 86.31 & 83.50 \\
S8 & 74.14 & 74.31 & 73.76 & 73.43 & 65.48 \\
S9 & 84.48 & 84.15 & 84.38 & 84.06 & 79.29 \\
\midrule
\textbf{Mean} & \textbf{68.20} & \textbf{69.67} & \textbf{68.60} & \textbf{67.40} & \textbf{57.33} \\
\textbf{Std}  & \textbf{12.13} & \textbf{11.05} & \textbf{11.29} & \textbf{11.99} & \textbf{16.22} \\
\bottomrule
\end{tabular}
\end{table}

\begin{table}[!t]
\caption{Cross-Subject Evaluation Results for ANFIS–FBCSP–PSO}
\label{tab:anfis_cross}
\centering
\footnotesize
\setlength{\tabcolsep}{3pt} 
\begin{tabular}{lccccc}
\toprule
\textbf{Subj.} & \textbf{Acc.} & \textbf{Prec.} & \textbf{Rec.} & \textbf{F1} & \textbf{Kappa} \\
\midrule
S1 & 75.86 & 74.90 & 75.16 & 74.76 & 67.56 \\
S2 & 63.79 & 67.46 & 64.93 & 62.97 & 52.37 \\
S3 & 81.03 & 80.94 & 80.24 & 79.96 & 74.70 \\
S4 & 62.07 & 62.50 & 63.36 & 61.90 & 49.57 \\
S5 & 43.10 & 43.68 & 43.08 & 41.90 & 24.17 \\
S6 & 48.28 & 53.76 & 45.22 & 44.94 & 26.11 \\
S7 & 84.48 & 83.62 & 86.63 & 84.67 & 78.91 \\
S8 & 77.59 & 78.65 & 81.00 & 79.23 & 69.71 \\
S9 & 55.17 & 55.46 & 54.78 & 54.13 & 39.85 \\
\midrule
\textbf{Mean} & \textbf{65.71} & \textbf{66.77} & \textbf{66.04} & \textbf{64.94} & \textbf{53.66} \\
\textbf{Std}  & \textbf{14.89} & \textbf{13.88} & \textbf{15.92} & \textbf{15.72} & \textbf{20.52} \\
\bottomrule
\end{tabular}
\end{table}
 
Tables~\ref{tab:summary_within} and~\ref{tab:summary_cross} summarize the mean and standard deviation of the performance metrics for both models in within-subject and cross-subject evaluations. These tables highlight the trade-off between subject-specific performance and cross-subject generalization.

\begin{table}[!t]
\caption{Within-Subject Classification Performance (Mean~$\pm$~Std) and Cohen’s Kappa of Benchmark Algorithms on the BCI IV-2a Dataset}
\label{tab:summary_within}
\centering
\footnotesize
\setlength{\tabcolsep}{6pt} 
\begin{tabular}{lcc}
\toprule
\textbf{Model} & \textbf{Accuracy (\%)} & \textbf{Kappa (\%)} \\
\midrule
EEGNet & 63.79~$\pm$~8.49 & 51.54~$\pm$~11.67 \\
ANFIS–FBCSP–PSO & 68.58~$\pm$~13.76 & 58.04~$\pm$~18.43 \\
\bottomrule
\end{tabular}
\end{table}

\begin{table}[!t]
\caption{Cross-Subject Classification Performance (Mean~$\pm$~Std) and Cohen’s Kappa of Benchmark Algorithms on the BCI IV-2a Dataset}
\label{tab:summary_cross}
\centering
\footnotesize
\setlength{\tabcolsep}{6pt} 
\begin{tabular}{lcc}
\toprule
\textbf{Model} & \textbf{Accuracy (\%)} & \textbf{Kappa (\%)} \\
\midrule
EEGNet & 68.20~$\pm$~12.13 & 57.33~$\pm$~16.22 \\
ANFIS–FBCSP–PSO & 65.71~$\pm$~14.89 & 53.66~$\pm$~20.52 \\
\bottomrule
\end{tabular}
\end{table}

\subsection{Statistical Significance Analysis}

To determine whether the observed performance differences between EEGNet and ANFIS–FBCSP–PSO were statistically significant, a paired Wilcoxon signed-rank test was performed on subject-level accuracy scores (N = 9) under each evaluation protocol. This non-parametric test accounts for paired observations by comparing the distribution of within-subject differences between the two models.

For the within-subject evaluation, the performance difference between ANFIS–FBCSP–PSO and EEGNet did not reach statistical significance ($p = 0.301$). Likewise, in the cross-subject (LOSO) evaluation, no statistically significant difference was observed between the two approaches ($p = 0.844$).

These results suggest that although mean performance trends differ across evaluation settings, the differences are not statistically significant given the limited sample size. Larger multi-subject datasets would be necessary to establish stronger statistical evidence.

\subsection{Discussion}

ANFIS–FBCSP–PSO demonstrated a higher mean performance trend in within-subject evaluation, benefiting from subject-specific feature extraction and fuzzy inference for personalized and interpretable modeling. However, this improvement was not statistically significant (p > 0.05), and should therefore be interpreted cautiously. The observed inter-subject variance (e.g., S7: 91.37\% vs S5: 48.27\%) likely reflects physiological and calibration differences, including variations in motor imagery ability, signal-to-noise ratio, and sensorimotor rhythm strength. Stronger and more consistent ERD/ERS patterns enhance CSP separability and favor ANFIS-based modeling, whereas noisier signals reduce performance.

In cross-subject (LOSO) evaluation, EEGNet showed a slightly higher mean generalization performance, though this difference was also not statistically significant (p > 0.05), indicating comparable overall performance under current dataset constraints.

A key advantage of ANFIS is its inherent interpretability through explicit fuzzy rules. For example:
\vspace{0.2cm}

\textit{IF} (Mu-band CSP feature is High) AND (Beta-band CSP feature is Medium) \\
\textit{THEN} Class = Right Hand.
\vspace{0.2cm}

Such rules provide physiologically meaningful insight into motor imagery decoding. Overall, ANFIS–FBCSP–PSO is well suited for personalized and explainable MI-BCI systems, while EEGNet offers a scalable end-to-end alternative with competitive cross-subject robustness. A limitation of this study is that real-time inference latency and deployment-level benchmarking were not explicitly recorded; future work will include systematic latency evaluation for practical MI-BCI deployment.

\section{Conclusion}
This work compared two paradigms for motor imagery based Brain Computer Interfaces (MI-BCIs): the interpretable ANFIS–FBCSP–PSO model and the deep learning-based EEGNet. ANFIS–FBCSP–PSO achieved higher within-subject accuracy through subject-specific feature extraction and fuzzy reasoning, while EEGNet generalized better across subjects due to its end-to-end spatial temporal learning. However, ANFIS is sensitive to inter-subject variability and parameter tuning, and EEGNet remains a black-box model with limited interpretability. These results emphasize that model choice should depend on the target application—EEGNet for robust generalization and ANFIS–FBCSP–PSO for interpretable, user-specific analysis. Future work will focus on hybrid and Transformer-based neuro-symbolic models that combine interpretability with scalability for practical, real-world MI-BCI deployment.

\bibliographystyle{IEEEtran}
\bibliography{ref.bib} 

\end{document}